\documentclass[sigconf]{acmart}

\newcommand{\methodname}{{\tt{FedGH}}}
\usepackage[ruled]{algorithm2e}

\newtheorem{assumption}{Assumption}[section]
\newtheorem{lemma}{Lemma}[section]
\newtheorem{theorem}{Theorem}
\usepackage{eucal}
\usepackage{arydshln} 
\usepackage{booktabs}
\usepackage{multirow}
\usepackage{subcaption}
\usepackage{caption}
\usepackage{amsfonts}
\usepackage{bbm}
\usepackage{xcolor}
\definecolor{ylp_color1}{RGB}{255,193,193}
\definecolor{ylp_color2}{RGB}{255,228,225}

\usepackage{pifont}
\usepackage{url}
\usepackage{balance}

\newcommand{\ReviewOne}[1]{\color{black}{#1}} 
\newcommand{\ReviewThree}[1]{\color{black}{#1}} 

\AtBeginDocument{%
  \providecommand\BibTeX{{%
    \normalfont B\kern-0.5em{\scshape i\kern-0.25em b}\kern-0.8em\TeX}}}

\copyrightyear{2023}
\acmYear{2023}
\setcopyright{acmlicensed}\acmConference[MM '23]{Proceedings of the 31st ACM International Conference on Multimedia}{October 29-November 3, 2023}{Ottawa, ON, Canada}
\acmBooktitle{Proceedings of the 31st ACM International Conference on Multimedia (MM '23), October 29-November 3, 2023, Ottawa, ON, Canada}
\acmPrice{15.00}
\acmDOI{10.1145/3581783.3611781}
\acmISBN{979-8-4007-0108-5/23/10}

\begin{document}

\title{FedGH: Heterogeneous Federated Learning with Generalized Global Header}

\author{Liping Yi}
\orcid{0000-0001-6236-3673}
\affiliation{%
  \institution{College of C.S., DISSec, GTIISC, Nankai University}
  \city{Tianjin}
  \country{China}
}
\email{yiliping@nbjl.nankai.edu.cn}

\author{Gang Wang}
\authornote{Corresponding authors.}
\orcid{0000-0003-0387-2501}
\affiliation{%
  \institution{College of C.S., DISSec, GTIISC, Nankai University}
  \city{Tianjin}
  \country{China}
}
\email{wgzwp@nbjl.nankai.edu.cn}

\author{Xiaoguang Liu}
\orcid{0000-0002-9010-3278}
\affiliation{%
  \institution{College of C.S., DISSec, GTIISC, Nankai University}
  \city{Tianjin}
  \country{China}
}
\email{liuxg@nbjl.nankai.edu.cn}

\author{Zhuan Shi}
\orcid{0000-0003-4239-3546}
\affiliation{%
  \institution{School of Computer Science and Technology, University of Science and Technology of China (USTC)}
  \city{Hefei}
  {Anhui}
  \country{China}
}
\email{zhuanshi@mail.ustc.edu.cn}

\author{Han Yu}
\authornotemark[1]
\orcid{0000-0001-6893-8650}
\affiliation{%
  \institution{School of Computer Science and Engineering, Nanyang Technological University (NTU)}
  \country{Singapore}
}
\email{han.yu@ntu.edu.sg}

\renewcommand{\shortauthors}{L. Yi and G. Wang, et al.}

\begin{abstract}
 Federated learning (FL) is an emerging machine learning paradigm that allows multiple parties to train a shared model collaboratively in a privacy-preserving manner. Existing horizontal FL methods generally assume that the FL server and clients hold the same model structure. However, due to system heterogeneity and the need for personalization, enabling clients to hold models with diverse structures has become an important direction. Existing model-heterogeneous FL approaches often require publicly available datasets and incur high communication and/or computational costs, which limit their performances. To address these limitations, we propose a simple but effective \underline{Fed}erated \underline{G}lobal prediction \underline{H}eader (\methodname{}) approach. It is a communication and computation-efficient model-heterogeneous FL framework which trains a shared generalized global prediction header with representations extracted by heterogeneous extractors for clients' models at the FL server. The trained generalized global prediction header learns from different clients. The acquired global knowledge is then transferred to clients to substitute each client's local prediction header. We derive the non-convex convergence rate of \methodname{}. Extensive experiments on two real-world datasets demonstrate that \methodname{} achieves significantly more advantageous performance in both model-homogeneous and -heterogeneous FL scenarios compared to seven state-of-the-art personalized FL models, beating the best-performing baseline by up to 8.87\% (for model-homogeneous FL) and 1.83\% (for model-heterogeneous FL) in terms of average test accuracy, while saving up to 85.53\% of communication overhead.
\end{abstract}




\maketitle

\section{Introduction}
Federated learning (FL) \citep{FL2019} has become a widely adopted approach for collaborative model training involving multiple participants with decentralized data under the premise of privacy preservation. Horizontal FL methods, such as {\tt{FedAvg}} \cite{FedAvg}, generally involve a central FL server coordinating multiple FL clients. In each round of distributed model training, the server broadcasts the global model to selected clients. The clients then train the received global model on their respective local datasets and send the updated local models back to the server. The server then updates the global model by aggregating the received local models. The above steps are iteratively executed until the global model converges. Since only the model parameters are transmitted between the server and clients without exposing the raw data, privacy protection is enhanced. Nevertheless, the above paradigm requires all clients to train models with the same structures (i.e., model homogeneity) in order to work.

However, in practical \textit{cross-device} FL scenarios, the clients participating in FL are mostly mobile edge devices with heterogeneous and constrained system resources (e.g., computing power, network bandwidth, memory, storage, and battery capacity) \cite{Yi-QSFL,Yu-1,Shi-2-FedFAIM,Shi-1-FedWM,Zhang-1,Zhang-2,Zhang-3}. This is also referred to as system heterogeneity in FL. Model-homogeneous FL methods face three limitations in this scenario: 
\begin{itemize}
\setlength{\itemsep}{1pt}
\setlength{\parskip}{1pt}
\setlength{\parsep}{1pt}
    \item \textbf{Device}: when training a large global model, some low-end clients may never be able to join in FL since their limited system resources preclude them from training large models. As a result, the accuracy of the final global model may be degraded due to the lack of information from these clients. 
    \item \textbf{Data}: the data held by different devices are often not identically and independently distributed (Non-IID), also known as statistical heterogeneity in FL \cite{Yu-2,Yu-3}.
    \item \textbf{Model}: if all clients join FL, the capacity of the trained homogeneous models must match the weakest client's system configurations. Unfortunately, training models with a small capacity not only reduces their performance but also wastes high-end clients' system resources due to long idle time. 
\end{itemize}
Although model-heterogeneous FL approaches have emerged to address the aforementioned challenges facing model-homogeneous FL, they still have the following limitations.
During learning, the high-level design intuition is to separate the training of the homogeneous portion and the heterogeneous portion of the FL model structure into unrelated processes. This not only results in limited performance improvement but also incurs high computation and communication costs \citep{LG-FedAvg, FML, FedKD}.
In addition, some approaches even rely on the availability of suitable public datasets closely related to the learning task in order to leverage knowledge distillation to achieve model-heterogeneous FL \citep{FedMD, FedDF}. However, this is not always viable in practice.
Therefore, enabling FL clients to train heterogeneous FL models with the capacity adaptive to system resource limitations and diverse data distributions in an efficient manner remains open.

To bridge the aforementioned gaps in the model-heterogeneous FL literature, we propose the \underline{Fed}erated \underline{G}lobal prediction \underline{H}eader (\methodname{}) approach. It is a novel model-heterogeneous FL framework capable of achieving low communication and computation costs. Under \methodname{}, each client's local model consists of a heterogeneous feature extractor and a homogeneous prediction header. It leverages the representations extracted by clients' feature extractors to train a global generalized prediction header at the server for all clients to share. The updated global header captures \textit{all-class} knowledge among multiple clients. The generalized global prediction header replaces each client's local prediction header to transfer global knowledge to clients. In this way, \methodname{} enables information interaction across heterogeneous clients' models through a shared generalized global prediction header. 

By communicating only the representations and the global prediction header's parameters between clients and the server, \methodname{} reduces communication costs. By computing local class-averaged representations on FL clients, it reduces computational costs to a level tolerable for mobile edge devices. By not relying on a public dataset, its operation is not limited by the availability of such datasets. By only sending representations which are high-level abstractions of local data, it protects data privacy. We prove the non-convex convergence rate of \methodname{}. Extensive experiments on two real-world datasets demonstrate that \methodname{} achieves significantly more advantageous performance in both model-homogeneous and -heterogeneous FL scenarios compared to seven state-of-the-art personalized FL models, beating the best-performing baseline by up to 8.87\% (for model-homogeneous FL) and 1.83\% (for model-heterogeneous FL) in terms of average test accuracy, while saving up to 85.53\% of communication overhead.


\section{Related Work}
Existing model-heterogeneous FL methods can be divided into two main categories: 1) each client's local model is a heterogeneous subnet of the server model, and 2) different clients hold completely heterogeneous local models. The former (such as {\tt{HeteroFL}}~\citep{HeteroFL}, {\tt{FjORD}}~\citep{FjORD}, {\tt{HFL}}~\citep{HFL}, {\tt{FedResCuE}}~\citep{FedResCuE}, {\tt{FedRolex}}~\citep{FedRolex} and {\tt{Fed2}}~\citep{Fed2}) allows clients to train heterogeneous subnets matching system resources to tackle system and statistical heterogeneity simultaneously, but the strong assumption of subnets constrains its applications. Our work is more closely related to the latter category, which can be further divided into two groups based on whether they rely on the availability of public datasets or not.

\textbf{Public Data-Dependent}. This category of methods achieves collaborative training across clients with heterogeneous models by knowledge distillation on public datasets. According to the site at which knowledge distillation is performed, these methods can be further divided into three groups.

\textit{Knowledge distillation on the clients}. In each communication round, {\tt{FedMD}}~\citep{FedMD} and {\tt{FSFL}}~\citep{FSFL} let clients compute the \textit{logits} of the trained local heterogeneous model on a public dataset, and uploads them to the server. The server then aggregates these logits to generate the global logits, and broadcasts them to clients. The clients calculate the distance between the local logits and the global logits belonging to one public data sample as the knowledge loss. Finally, the distilled local model is fine-tuned on private data. To speed up convergence or enhance robustness to adversarial attacks of the above approach, {\tt{Cronus}}~\citep{Cronus}, {\tt{DS-FL}}~\citep{DS-FL} and {\tt{FedAUX}}~\citep{FEDAUX} proposed new aggregation rules for logits. Instead of communicating logits, {\tt{FedHeNN}}~\citep{FedHeNN} extracts \textit{representations} in the above distillation process. 

\textit{Knowledge distillation on server}. {\tt{FedDF}}~\citep{FedDF}, {\tt{FCCL}}~\citep{FCCL}, {\tt{FedKT}}~\citep{FedKT}, {\tt{Fed-ET}}~\citep{Fed-ET} and {\tt{FedKEMF}}~\citep{FedKEMF} train each client's heterogeneous model via ensemble distillation on a public dataset at the server. 

\textit{Knowledge distillation on both the clients and the server}. Upon distillation at the client side, {\tt{FedGEMS}}~\citep{FedGEMS} and {\tt{CFD}}~\citep{CFD} include one additional step of distillation on the server's model to mitigate forgetfulness due to dropout. 

However, the public datasets essential for the above approaches to work may not always be available in practice. Furthermore, only public data following similar distributions with clients' private data can obtain acceptable model performance, which makes them even harder to find. Besides, distillation on each sample of public data incurs non-trivial computation costs if the public data size is large. These facts limit the applicability of these approaches.

\textbf{Public Data-Independent}. It involves three lines: model mixup, mutual learning and data-free knowledge distillation.

\textit{Model mixup}: there are many studies that split each client's local model into two parts: a feature extractor and a classifier. Only one part is shared during FL model aggregation, while the other part containing personalized parameters or even heterogeneous structures is held locally.
{\tt{FedRep}}~\citep{FedRep}, {\tt{FedMatch}}~\citep{FedMatch}, {\tt{FedBABU}}~\citep{FedBABU} and {\tt{FedAlt/FedSim}}~\citep{FedAlt/FedSim} share the homogeneous feature extractor while {\tt{LG-FedAvg}}~\citep{LG-FedAvg}, {\tt{CHFL}}~\citep{CHFL} and {\tt{FedClassAvg}}~\citep{FedClassAvg} share the homogeneous classifier header. Since only part of a complete model is shared, model performance tends to degrade compared with sharing the complete model (e.g., {\tt{FedAvg}}). Besides, a feature extractor has more parameters than a classifier header. Thus, allowing different clients to use heterogeneous extractors boosts FL model heterogeneity. Hence, we choose to allow clients to hold personalized heterogeneous feature extractors and share their homogeneous classifier headers via FL global training.

\textit{Mutual learning}: {\tt{FML}}~\citep{FML} and {\tt{FedKD}}~\citep{FedKD} enable each client to train a large heterogeneous model and a small homogeneous model via mutual learning, and the small homogeneous models are aggregated on the server. Since each client is required to train two models simultaneously, the extra computation overhead may not be tolerable for mobile edge devices. 

\textit{Data-free knowledge distillation}: {\tt{FedGen}}~\citep{FedGen} trains a generator with clients' local data distribution on the server to learn the overall distribution. The trained generator produces extra representation with the overall distribution for each client to enhance local model generalization. However, uploading local data distributions from clients to the server risks exposing data privacy. In {\tt{FedZKT}}~\citep{FedZKT}, the server trains a generative model and a global model in an adversarial manner to transfer local knowledge to the global model. It uses the trained generative model to produce synthetic data for distilling the global knowledge to local models. The computation-intensive adversarial training and knowledge distillation are time-consuming. {\tt{FedGKT}}~\citep{FedGKT} communicates features, logits and labels of clients' local data with the server to distil small clients' classifiers and a large server's classifier bidirectionally. Since the server and clients exchange information for each private sample, the communication cost is high when the private dataset is large. {\tt{FD}}~\citep{FD} aggregates logits by class on the server, and clients calculate the distance of each local sample logits and the aggregated global logits as distillation loss to train local models. Since logits carry similar information with hard labels, no extra knowledge is supplemented, which tends to degrade performance. To improve {\tt{FD}}, {\tt{HFD}}~\citep{HFD1, HFD2} allows clients to upload averaged samples by class, which increases the risk of privacy leakage. Different from {\tt{FD}}, {\tt{FedProto}}~\citep{FedProto} utilizes representations rather than logits by class. The server in {\tt{FedProto}} aggregates the received representations with \textit{class distributions} as weights instead of averaging the received logits like {\tt{FD}}. This potentially risks privacy leakage. Both {\tt{FD}} and {\tt{FedProto}} need to compute the distillation loss between each private sample logits/representations and global logits/representations with the corresponding class, which incurs high computation costs at client sides. In addition, each client can only learn about classes it already knows from the server, which hinders generalization to unseen classes.

Unlike {\tt{FedProto}}, \methodname{} utilizes local representations and the corresponding \textit{classes (labels)}, rather than \textit{class distributions}, to train a homogeneous shared global prediction header at the server, and then uses it to replace local model headers to achieve global knowledge transfer. The shared global header captures all-class information across different clients whose local models consist of heterogeneous extractors and homogeneous prediction headers, thereby enhancing the generalization of local models. By not requiring \textit{class distributions}, \methodname{} reduces privacy leakage. {\tt{LG-FedAvg}} directly aggregates homogeneous local headers on the server, which can also support heterogeneous clients' extractors. However, the simple weighted averaging of headers by data size is ineffective in the face of non-IID data. In \methodname{}, each client only provides the local averaged representation (one embedding vector) about each seen class to train a global generalized header, which can better accommodate non-IID data. 

\section{The Proposed \methodname{} Approach}
In this section, we first describe the formulation of a typical FL algorithm - {\tt{FedAvg}}, and then define the problem \methodname{} addresses. We then explain how \methodname{} works for model-heterogeneous FL, and discuss its strengths in cost reduction and privacy preservation.

\subsection{Preliminaries}
\textbf{Typical FL}. Under {\tt{FedAvg}}, a central FL server coordinates $N$ FL clients to collaboratively train a global model. Specifically, in each training round $t$, the server samples a fraction of all the clients, $C$, to join training (i.e., the set of sampled clients joining in the $t$-th round of FL, $|\mathcal{S}^t|=C \cdot N =K$). Then, the server broadcasts the global model $\omega$ to the $K$ selected clients. They then train the received global model on their respective local data $D_k\sim P_k$ ($D_k$ obeys the distribution $P_k$, i.e., the local data of different clients are non-IID) to obtain $\omega_k$ through $\omega_k \leftarrow \omega-\eta \nabla \ell\left(\omega;\boldsymbol{x}_i, y_i\right),\left(\boldsymbol{x}_i, y_i\right) \in D_k$. The $k$-th client uploads the trained local model $\omega_k$ to the server. The server then aggregates them to update the global model as $\omega=\sum_{k=0}^{K-1} \frac{n_k}{n} \omega_k$. In short, {\tt{FedAvg}} aims to minimize the average loss of the global model $\omega$ on all clients' local data:
\begin{equation}
\min _{\omega \in \mathbb{R}^d} \mathcal{L}(\omega)=\sum_{k=0}^{K-1} \frac{n_k}{n} \mathcal{L}_k(\omega),
\end{equation}
where $n_k=|D_k|$ is the number of samples held by the $k$-th client. $n$ is the number of samples held by all clients. $\mathcal{L}_k(\omega)=\ell\left( \omega;D_k\right)$ is the loss of the global model $\omega$ with $d$ dimensions on the $k$-th client's local data $D_k$.

The above steps iterate until the global model converges. Since the server averages the received local models, the structures of all clients' local models must be the same (homogeneous).

\textbf{Problem Definition for \methodname{}}. We aim to perform FL across clients with heterogeneous models in the same supervised classification tasks. Each client's local model can be split into two parts: $f\left(\omega_k\right)=\mathcal{F}_k\left(\varphi_k\right) \circ \mathcal{H}_k\left(\theta_k\right)$, i.e., $\omega_k=\left(\varphi_k, \theta_k\right)$, where $\circ$ denotes model splicing. $\mathcal{F}_k\left(\varphi_k;\boldsymbol{x}\right)$: $\mathbb{R}^{d_{\boldsymbol{x}}} \rightarrow \mathbb{R}^{d_{\boldsymbol{\mathcal{R}}}}$ is a feature extractor, which maps local samples from the input feature $\boldsymbol{x}$ to the representation embedding $\boldsymbol{\mathcal{R}}$. $\mathcal{H}_k\left(\theta_k; \mathcal{F}_k\left(\varphi_k;\boldsymbol{x}\right)\right)$: $\mathbb{R}^{d_{\boldsymbol{\mathcal{R}}}} \rightarrow \mathbb{R}^{d_y}$ is the prediction header. All clients have the same $d_{\boldsymbol{x}}$, $d_{\boldsymbol{\mathcal{R}}}$, $d_y$. We assume that $\mathcal{F}_k$ is heterogeneous across different clients (i.e., clients can customize the sizes and structures of local feature extractors to match their system resources and data volume), and all clients share the homogeneous global header $\mathcal{H}(\theta)$ (i.e., all clients carry out the same tasks). That is, $f\left(\omega_k\right)=\mathcal{F}_k\left(\varphi_k\right) \circ \mathcal{H}(\theta)$. So the loss of the $k$-th client's local model is formulated as $\mathcal{L}_k\left(\omega_k ; \boldsymbol{x}, y\right)=\mathcal{L}_{\text {sup }}\left(\mathcal{H}\left(\theta; \mathcal{F}_k\left(\varphi_k ; \boldsymbol{x}\right)\right), y\right),(\boldsymbol{x}, y) \in D_k$.

In representation learning \citep{RepLearning}, representations are the latent feature embedding vectors extracted by feature extractors from input samples. It is hard to infer the original data from the representations without knowing the model parameters \citep{FedProto}. Therefore, we utilize the representations with the same dimension extracted by different clients' heterogeneous feature extractors and the corresponding labels (classes) to train a shared global prediction header on the server. It acquires knowledge across all clients and all classes. Clients with homogeneous local models are the special cases of this scenario. We define the training goal of \methodname{} as minimizing the sum of the losses of all clients' local heterogeneous models $\{\omega_0, \ldots, \omega_{N-1}\}$ with dimensions $\{d_0, \ldots, d_{N-1}\}$:
\begin{equation}
\min _{\omega_0, \ldots, \omega_{N-1} \in \mathbb{R}^{d_0, \ldots, d_{N-1}}} \sum_{k=0}^{N-1} \mathcal{L}_k\left(\omega_k\right), \omega_k=\varphi_k \circ \theta.
\end{equation}

\subsection{Federated Global Header (\methodname{}) Algorithm}

\begin{figure}[!t]
  \centering
  \includegraphics[width=1\linewidth]{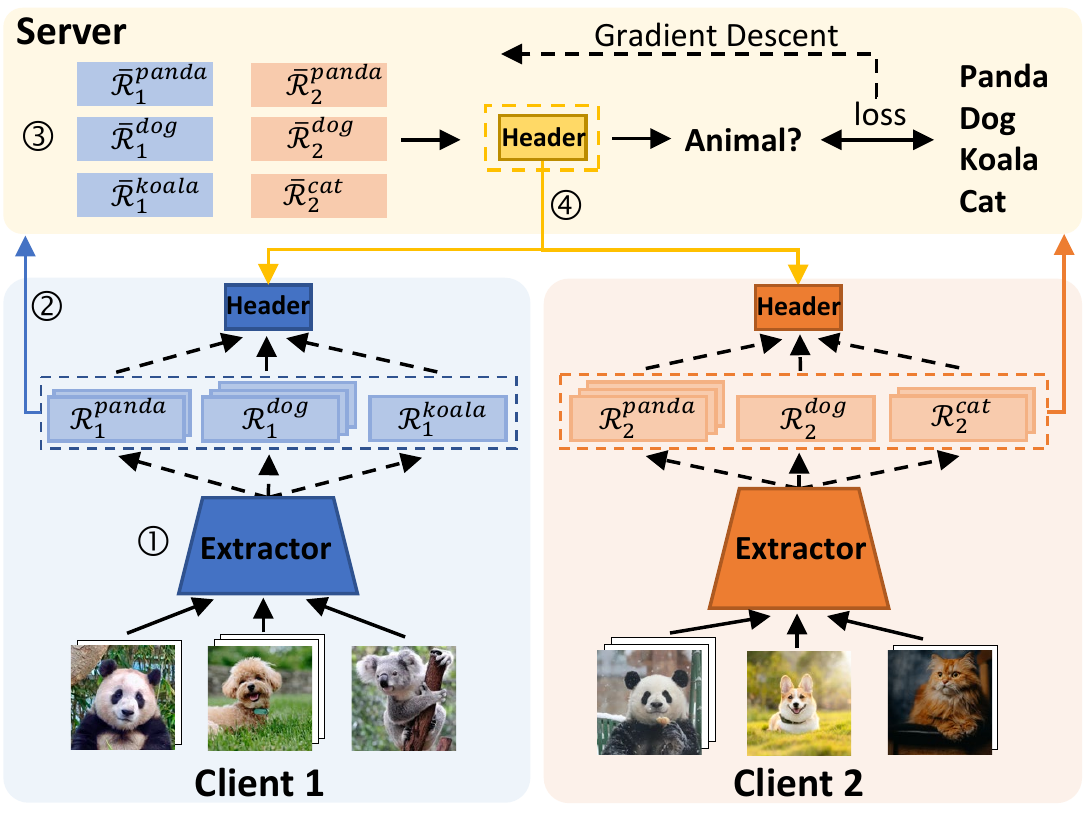}
  \caption{The workflow of the proposed \methodname{} approach. In each communication round: \ding{192} Clients train local heterogeneous models on local data. \ding{193} Clients' feature extractors output representations of all local data samples and calculate the average of representations belonging to the same class. Then, the local averaged representation and label for each class are uploaded to the server; \ding{194} The server uses the received local-averaged representations and class labels to train the global prediction header, then broadcasts it to the clients. \ding{195} Clients replace their local prediction header with the received shared global header. Steps \ding{192}-\ding{195} are repeated until all clients' local models converge. After federated training, heterogeneous local models are used for inference.}
  \label{fig:FedGH}
\end{figure}

The workflow of \methodname{} is displayed in Figure \ref{fig:FedGH}. In the $t$-th FL training round, the $k$-th client uses its feature extractor $\varphi_k^t$ of the local heterogeneous model $\omega_k^t$ after local training to extract the representations $\boldsymbol{\mathcal{R}}_{k, i}^t$ of each local training sample $\left(\boldsymbol{x}_i, y_i\right)$ in $D_k$. Then, it calculates the average representation of samples within the same class $s$ as the local averaged representation $\overline{\boldsymbol{\mathcal{R}}}_k^{t, s}$ (abbr. LAR) of the corresponding class:
\begin{equation}
\overline{\boldsymbol{\mathcal{R}}}_k^{t, s}=\frac{1}{\left|D_k^s\right|} \sum_{i \in D_k^s} \boldsymbol{\mathcal{R}}_{k, i}^t=\frac{1}{\left|D_k^s\right|} \sum_{i \in D_k^s} \mathcal{F}_k\left(\varphi_k^t;\boldsymbol{x}_i\right).
\end{equation}

The $k$-th client uploads the LARs $\overline{\boldsymbol{\mathcal{R}}}_k^{t, s}$ for each of its local classes and the corresponding class label $s$ to the server. As stated in \citet{FedProto}, the representations are latent feature embedding vectors extracted from the data. Thus, it is hard to infer original data inversely with only extracted representations and without the parameters of the feature extractors. Since each client uploads LARs (i.e., class-wise averaged representations), the risk of privacy leakage is reduced further.


The server inputs all the received LARs $\overline{\boldsymbol{\mathcal{R}}}_k^{t, s}$ from $K$ participating clients into the global prediction header $\mathcal{H}$ to produce the prediction. The hard loss (e.g., cross-entropy loss) between the output prediction and the true class label $s$ is used to update the global header parameters $\theta^{t-1}$ via gradient descent:
\begin{equation}
\theta^t \leftarrow \theta^{t-1}-\eta_{\mathcal{\theta}} \nabla \ell\left(\theta^{t-1}; \overline{\boldsymbol{\mathcal{R}}}_k^{t, s}, s\right),
\end{equation}
where $\eta_{\mathcal{\theta}}$ is the learning rate of the global prediction header. To improve the efficiency of training the global prediction header, we allow the server to train the global header once a client's LARs are received. After the LARs from all participating clients are fed into the global header for training, the global prediction header is updated in the current round.
The updated global header acquires all-class knowledge across different clients. Thus, it has a stronger generalization capability than local headers with partial-class knowledge.

The server broadcasts the updated global header $\theta^{t}$ to the clients selected for the next training round. In the $(t+1)$-th round, the $k$-th client replaces its local prediction header $\theta_{k}^{t}$ with the received global header $\theta^{t}$. In this way, its complete local model becomes:
\begin{equation}
\widetilde{\omega}_k^{t+1}=\varphi_k^t \circ \theta^t.
\end{equation}

Intuitively, clients' local models can converge faster with the generalized global header. Besides, the spliced complete local model obtains the old local knowledge from the personalized heterogeneous feature extractor and the new global knowledge from the shared global header, which enables it to better deal with statistical heterogeneity.

The assembled complete local model $\widetilde{\omega}_k^{t+1}$ is trained on local data $D_k$ to obtain the updated local model ${\omega}_k^{t+1}$:
\begin{equation}
\omega_k^{t+1} \leftarrow \widetilde{\omega}_k^{t+1}-\eta_\omega \nabla \ell\left(\widetilde{\omega}_k^{t+1};D_k\right),
\end{equation}
where $\eta_\omega$ is the local model learning rate. 

The above steps iterate until all local heterogeneous models converge. The pseudocode for \methodname{} can be found in Algorithm~\ref{alg:FedGH}.

\subsection{Discussion}
Here, we analyze the strength of \methodname{} in cost reduction and privacy preservation.

\textbf{Computation Cost}. Under \methodname{}, clients are required to compute the representation for each local training data sample and the averaged representation for samples belonging to the same class. Extracting the representation for one sample is a forward inference of the local model on this sample. Thus, extracting representations only consumes half the computation cost of local training (forward and backwards) in one epoch. Generally, the epochs of local training are set to be larger than 1 in order to avoid frequent communications during FL model training \cite{FedAvg}. Therefore, extracting representations consumes acceptable computation cost. Besides, since one representation is an $r\times1$ vector, to calculate the average of representations belonging to each class held by a client, we can first use a ``variant'' to stack the sum of the representations of each class, and then calculate the average. Therefore, when calculating local average representation (LAR), each client incurs a storage cost and the computational complexity is $\mathcal{O}(n)$, which are negligible compared to the cost of local model training.


On the server side, the computation cost of using LARs to train a shared global header is much lower than training a complete model as the global header is part of a complete model and the number of LARs is far fewer than local data samples. Besides, since the server often has sufficient computation power, training a global header consumes an acceptable portion of its computation resources. 

Overall, due to negligible computation cost on both the client and server, \methodname{} is suitable for both cross-device FL scenarios with resource-constrained mobile edge devices and cross-silo FL scenarios with more powerful participants.

\textbf{Communication Cost}. During the \textit{client-to-server uplink} communication, clients upload the LAR and the class label for each class to the server. The class label is an integer-type value and the LAR is an $r\times 1$ vector. If each client has $S$ classes, \methodname{} incurs $(S+S\times r) \times 32$~bits of communication cost, which can be negligible compared to uploading the complete local model in {\tt{FedAvg}}. 

During \textit{server-to-client downlink} communication, the server broadcasts the updated global header parameters to clients. This incurs lower communication costs than broadcasting the complete global model to clients in {\tt{FedAvg}}. Thus, \methodname{} is communication-efficient. 

\textbf{Privacy Preservation}. During the \textit{client-to-server uplink} communication, clients upload the LAR and the class label for each class to the server. As stated above, the representation for a sample is an embedding vector mapped by the feature extractor from the original feature space to the embedding space. Thus, it is hard to infer the original data by stealing only representations without knowing the parameters of the feature extractor. Moreover, the uploaded LAR is a mixup of representations within the same class, which further enhances privacy protection. 

During the \textit{server-to-client downlink} communication, the server broadcasts the global prediction header to clients. Since it is part of a complete model, it is also difficult to infer original data by just knowing the global prediction header. Hence, \methodname{} achieves a high level of privacy preservation. It can be combined with existing privacy protection mechanisms to further enhance FL security.

\begin{algorithm}[t]
  \caption{FedGH}
  \label{alg:FedGH}
  \textbf{Input}: $N$, total number of clients; $K$, number of selected clients in one round; $T$, number of rounds; $\eta_{\omega}$, learning rate of local models; $\eta_{\theta}$, learning rate of global header. \\
     Randomly initialize the heterogeneous local models $\left[\omega_0^0, \ldots, \omega_{N-1}^0\right]$ and global header $\theta^0$. \\
    \For{$t=0$ {\bfseries to} $T-1$}{
         $\mathcal{S}^t \leftarrow$  Randomly select $K\leqslant N$ clients to join FL. \\
         \vspace{1em}
         // \textbf{Clients Side} (each client $k\in\mathcal{S}^t$): \\
                  Receive the global header $\theta^{t-1}$ broadcast by the server; \\
                  Update the local model: $\widetilde{\omega}_k^t=\varphi_k^{t-1} \circ \theta^{t-1}$; \\
                  Perform local training: $\omega_k^t \leftarrow \widetilde{\omega}_k^t-\eta_\omega \nabla \ell\left(\widetilde{\omega}_k^t ; D_k\right)$; \\
                  Calculate the representation $\boldsymbol{\mathcal{R}}_{k, i}^t$ of each private training sample $i \in D_k$ on the trained local model $\omega_k^t$; \\
                  Calculate the average representation for each local class: 
                  $\overline{\boldsymbol{\mathcal{R}}}_k^{t, s}=\frac{1}{\left|D_k^s\right|} \sum_{i \in D_k^s} \boldsymbol{\mathcal{R}}_{k, i}^t=\frac{1}{\left|D_k^s\right|} \sum_{i \in D_k^s} \mathcal{F}_k^t\left(\varphi_k^{t};\boldsymbol{x}_i\right)$; \\
                  Upload each averaged local class representation $\overline{\boldsymbol{\mathcal{R}}}_k^{t, s}$ and the corresponding class label $s$ to the server. \\
  \vspace{1em}
           // \textbf{Server Side}: \\
           Receive the averaged local class representation $\overline{\boldsymbol{\mathcal{R}}}_k^{t, s}$ and corresponding class label $s$ from the selected $K$ clients; \\
           // Train the global header: \\
          \For{$k\in\mathcal{S}^t$}{
               $\theta^t \leftarrow \theta^{t-1}-\eta_\theta \nabla \ell\left(\theta^{t-1} ; \overline{\boldsymbol{\mathcal{R}}}_k^{t, s}, s\right)$; 
          }
           Broadcast the trained global header $\theta^{t}$ to the clients selected in the next round of training. \\
    }
     \textbf{Return} Personalized heterogeneous private models for all clients:  $\left[\omega_0^{T-1}, \omega_1^{T-1}, \ldots, \omega_{N-1}^{T-1}\right]$.
\end{algorithm}

\section{Convergence Analysis}
To analyze the convergence of \methodname{}, we first introduce some additional notations. $t$ indicates the current communication round, $e \in \{0, 1, ..., E\}$ is a local iteration, with up to $E$ iterations being executed. $(tE+e)$ is the $e$-th iteration in the $(t+1)$-th round. $(tE+0)$ indicates that at the beginning of the $(t+1)$-th round, clients replace their local prediction header with the global header trained in the $t$-th round. $(tE+1)$ is the first iteration in the $(t+1)$-th round. $(tE+E)$ denotes the last iteration in the $(t+1)$-th round. 

\begin{assumption}\label{assump:Lipschitz}
\textbf{Lipschitz Smoothness}. The $k$-th client's local model gradient is $L1$--Lipschitz smooth, i.e.,
\begin{equation}\label{eq:7}
\begin{gathered}
\left\|\nabla \mathcal{L}_k^{t_1}\left(\omega_k^{t_1} ; \boldsymbol{x}, y\right)-\nabla \mathcal{L}_k^{t_2}\left(\omega_k^{t_2} ; \boldsymbol{x}, y\right)\right\| \leqslant L_1\left\|\omega_k^{t_1}-\omega_k^{t_2}\right\|, \\
\forall t_1, t_2>0, k \in\{0,1, \ldots, N-1\},(\boldsymbol{x}, y) \in D_k.
\end{gathered}
\end{equation}

From Eq. \eqref{eq:7}, we can further derive:
\begin{equation}
\mathcal{L}_k^{t_1}-\mathcal{L}_k^{t_2} \leqslant\left\langle\nabla \mathcal{L}_k^{t_2},\left(\omega_k^{t_1}-\omega_k^{t_2}\right)\right\rangle+\frac{L_1}{2}\left\|\omega_k^{t_1}-\omega_k^{t_2}\right\|_2^2 .
\end{equation}

\end{assumption}

\begin{assumption} \label{assump:Unbiased}
\textbf{Unbiased Gradient and Bounded Variance}. The random gradient $g_k^t=\nabla \mathcal{L}_k^t\left(\omega_k^t ; \mathcal{B}_k^t\right)$ ($\mathcal{B}$ is a batch of local data) of each client's local model is unbiased, i.e.,
\begin{equation}
\mathbb{E}_{\mathcal{B}_k^t \subseteq D_k}\left[g_k^t\right]=\nabla \mathcal{L}_k^t\left(\omega_k^t\right),
\end{equation}
and the variance of random gradient $g_k^t$ is bounded by:
\begin{equation}
\mathbb{E}_{\mathcal{B}_k^t \subseteq D_k}\left[\left\|\nabla \mathcal{L}_k^t\left(\omega_k^t ; \mathcal{B}_k^t\right)-\nabla \mathcal{L}_k^t\left(\omega_k^t\right)\right\|_2^2\right] \leqslant \sigma^2 .
\end{equation}    

\end{assumption} 

\begin{assumption} \label{assump:header}
\textbf{Bounded Variance of the Prediction Header}. The variance of the local prediction header $\mathcal{H}_k\left(\theta_k\right)$ for the local model $\omega_k$ trained on the client $k$'s local data $D_k$, and the global prediction header $\mathcal{H}(\theta)$ trained on the global data indirectly through LAR are bounded, i.e.,

\textit{parameter bounded}: $\mathbb{E}\left[\left\|\theta_k-\theta\right\|_2^2\right] \leqslant \varepsilon^2$,

\textit{gradient bounded}: $\mathbb{E}\left[\left\|\nabla \mathcal{L}\left(\theta_k\right)-\nabla \mathcal{L}(\theta)\right\|_2^2\right] \leqslant \delta^2$.    
\end{assumption}

Based on the above assumptions, since \methodname{} makes no change to the local model training process, Lemma \ref{lemma:LocalTraining} derived by \citet{FedProto} still holds.

\begin{lemma} \label{lemma:LocalTraining}
Based on Assumptions~\ref{assump:Lipschitz} and \ref{assump:Unbiased}, during the $\{0,1,...,E\}$ local iterations of the $(t+1)$-th FL training round, the loss of an arbitrary client's local model is bounded by:
\begin{equation}
\mathbb{E}\left[\mathcal{L}_{(t+1) E}\right] \leqslant \mathcal{L}_{t E+0}-\left(\eta-\frac{L_1 \eta^2}{2}\right) \sum_{e=0}^E\left\|\mathcal{L}_{t E+e}\right\|_2^2+\frac{L_1 E \eta^2}{2} \sigma^2 .
\end{equation}
\end{lemma}

\begin{lemma} \label{lemma:AfterAggregation}
Based on Assumption~\ref{assump:header}, the loss of an arbitrary client's local model (the local prediction header of which is replaced with the latest global prediction header) is bounded by:
\begin{equation}
\mathbb{E}\left[\mathcal{L}_{(t+1) E+0}\right] \leqslant \mathbb{E}\left[\mathcal{L}_{(t+1) E}\right]+\frac{\eta L_1 \delta^2}{2}.
\end{equation}
\end{lemma}
The detailed proof can be found in Appendix~\ref{sec:proof-Lemma2}.

Based on Lemma~\ref{lemma:LocalTraining} and Lemma \ref{lemma:AfterAggregation}, we can further derive the following theorems.

\begin{theorem} \label{theorem:one-round}
\textbf{One-round deviation}. Based on the above assumptions, the expectation of the loss of an arbitrary client's local model before the start of a round of local iteration satisfies
\begin{equation}
\begin{aligned}
\mathbb{E}\left[\mathcal{L}_{(t+1) E+0}\right] \leqslant & \mathcal{L}_{t E+0}-\left(\eta-\frac{L_1 \eta^2}{2}\right) \sum_{e=0}^E\left\|\mathcal{L}_{t E+e}\right\|_2^2 \\
& +\frac{\eta L_1\left(E \eta \sigma^2+\delta^2\right)}{2}.
\end{aligned}
\end{equation}
\end{theorem}

The proof can be found in Appendix~\ref{sec:proof-Theorem1}.

\begin{theorem} \label{theorem:non-convex}
\textbf{Non-convex convergence rate of {\ReviewOne{FedGH}}}. The above assumptions, for an arbitrary client and any $\epsilon>0$, the following inequality holds:
\begin{equation}
\begin{aligned}
\frac{1}{T} \sum_{t=0}^{T-1} \sum_{e=0}^E \mathbb{E}\left[\left\|\mathcal{L}_{t E+e}\right\|_2^2\right] & \leqslant  \frac{2\left(\mathcal{L}_{t=0}-\mathcal{L}^*\right)}{T \eta\left(2-L_1 \eta\right)}+\frac{L_1\left(E \eta \sigma^2+\delta^2\right)}{2-L_1 \eta} \\
& \leqslant \epsilon, \\
\text { s.t. } \eta & < \frac{2 \epsilon-L_1 \delta^2}{L_1\left(\epsilon+E \sigma^2\right)}.
\end{aligned}
\end{equation}
\end{theorem}
Therefore, under \methodname{}, an arbitrary client's local model can converge at the non-convex convergence rate $\epsilon \sim \mathcal{O}\left(\frac{1}{T}\right)$. The detailed proof can be found in Appendix~\ref{sec:proof-Theorem2}.

\section{Experimental Evaluation}
In this section, we experimentally compare \methodname{} \footnote{\url{https://github.com/LipingYi/FedGH}} with seven existing approaches on two real-world datasets.
We implement \methodname{} and all baselines with PyTorch and simulate the FL processes on NVIDIA GeForce RTX 3090 GPUs with 24G memory. 

\subsection{Experiment Setup}
\textbf{Datasets and Models}. We evaluate \methodname{} and baselines on two image classification datasets: CIFAR-10 and CIFAR-100 \footnote{\url{https://www.cs.toronto.edu/\%7Ekriz/cifar.html}} \citep{cifar}, which are manually divided into non-IID datasets following the method in \citet{pFedHN}. Specifically, for CIFAR-10, we assign only data from 2 out of the 10 classes to each client (non-IID: 2/10). For CIFAR-100, we assign only data from 10 out of the 100 classes to each client (non-IID: 10/100). Then, each client's local data are further divided into the training set, the evaluation set and the testing set following the ratio of 8:1:1. In this way, the testing set is stored locally by each client which follows the same distribution as the local training set. For the CIFAR-10 and CIFAR-100 datasets, each client trains a CNN model and a ResNet-18 model, respectively. The dimensions of the output layer (i.e., the last fully-connected layer) are $10$ and $100$, and the dimensions of the representation layer (i.e., the second last layer) are set to be $500$.

\textbf{Baselines}. We compare \methodname{} with the following methods. {\tt{Standalone}}, each client trains its local model independently, which serves as a lower bound of model performance. {\tt{FedAvg}} \citep{FedAvg}, a popular FL algorithm that only supports homogeneous local models. The public-data independent model-heterogeneous FL methods include {\tt{FML}} \citep{FML}, {\tt{FedKD}} \citep{FedKD} with mutual learning, {\tt{LG-FedAvg}} \citep{LG-FedAvg} with model mixup, {\tt{FD}} \citep{FD} with knowledge distillation on logits within the same class, and {\tt{FedProto}} \citep{FedProto} with knowledge distillation on representations within the same class.

\textbf{Evaluation Metrics}. \textbf{Accuracy}: we measure the accuracy ($\%$) of each client's local model and report the average test accuracy of all clients' local models. \textbf{Communication Overhead (CO)}: We record the communication overhead ($MB$) incurred upto the point in time when the FL model reaches the target accuracy, which is calculated as (number of rounds required $\times$ the number of clients in each round $\times$ number of floating point data transmitted in the uplink and downlink per round per client $\times$ $32$ bits).

\textbf{Training Strategy}. We tune optimal FL settings for all methods via grid search. The epochs of local training: $E \in \{1, 10, 30, 50, 100\}$ and the batch size of local training: $B \in \{32, 64, 128, 256, 512\}$. The optimizer of local training is SGD with learning rate $\eta_\omega=0.01$. We also tune special hyperparameters for baselines and report the optimal results. Note that \methodname{} introduces no additional hyperparameters except the global prediction header learning rate $\eta_\theta$. We set $\eta_\theta = \eta_\omega = 0.01$ by default. To compare \methodname{} with the baselines fairly, we set the total number of communication rounds $T \in \{100, 500\}$ to guarantee that all algorithms converge.

{\ReviewOne{
\textbf{Training process of \methodname{}}.
\textbf{Client:} On the CIFAR-10 (non-IID: 2/10) dataset, each client uses its local heterogeneous feature extractor after local training with learning rate  $\eta_\omega = 0.01$ to extract the representation embedding of each data sample and compute the local averaged representation (LAR) for each class. Then each client uploads LARs and labels of its held 2 classes to the server. Similarly, on CIFAR-100 (non-IID: 10/100) dataset, each client sends 10 classes’ LARs and labels to the server. \textbf{Server:} In the order of client id, the server inputs the LAR of a class from one client into the global header once, then computes the hard loss between the global header output and the label to update the global header via gradient descent with a learning rate $\eta_\theta = 0.01$. After LARs and labels from all participating clients have been processed, the global header updating in a given round is finished. Furthermore, to accelerate training the global header, we can regard the LARs and the corresponding labels from one client as a batch and allow the server to execute mini-batch gradient descent, which is necessary for the FL scenarios with a large number of clients or classes held by each client.
}}

\subsection{Results and Discussion}
Model-homogeneous FL can be regarded as a special case of model-heterogeneous FL. Thus, we first evaluate the approaches under the model-homogeneous FL setting before evaluating them under the model-heterogeneous FL setting.

\begin{table}[t]
\centering
\caption{Comparison of average test accuracy ($\%$) under the model-homogeneous FL setting, with different total numbers of clients $N$ and client participating rate $C$. ``-'' indicates that the algorithm fails to converge.}
\vspace{-0.5em}
\resizebox{1\linewidth}{!}{
\begin{tabular}{|l|c|c|c|c|c|c|}
\hline
             & \multicolumn{2}{c|}{$N=10, C=100\%$} & \multicolumn{2}{c|}{$N=50, C=20\%$} & \multicolumn{2}{c|}{$N=100, C=10\%$} \\ \cline{2-7} 
Method       & CIFAR-10        & CIFAR-100      & CIFAR-10         & CIFAR-100       & CIFAR-10         & CIFAR-100        \\ \hline
{\tt{Standalone}}   & 93.13           & 62.80          & 95.39            & 62.38           & 92.92            & 55.47            \\
{\tt{FedAvg}}      & 94.34           & 64.63          & 95.68            & 62.95           & 93.39            & 56.23            \\ 
{\tt{FML}}         & 92.39           & 61.58          & 94.55            & 56.80           & 90.36            & 50.16            \\
{\tt{FedKD}}        & 92.65           & 58.35          & 93.93            & 57.36           & 91.07            & 51.90            \\
{\tt{LG-FedAvg}}    & 93.54           & 63.30          & 95.29            & 63.06           & 92.96            & 54.89            \\
{\tt{FD}}         & 93.63           & -              & -                & -               & -                & -                \\
{\tt{FedProto}}    & 95.99           & 62.51          & 95.38            & 61.15           & 92.75            & 55.53            \\ \hline
\methodname{} & \textbf{96.33}  & \textbf{73.62} & \textbf{95.69}   & \textbf{65.02}  & \textbf{93.65}   & \textbf{56.44}   \\ \hline
\end{tabular}
\label{tab:model-homo}
}
\vspace{-0.5em}
\end{table}

\begin{figure}[t]
\centering
\begin{minipage}[t]{0.5\linewidth}
\centering
\includegraphics[width=1.8in]{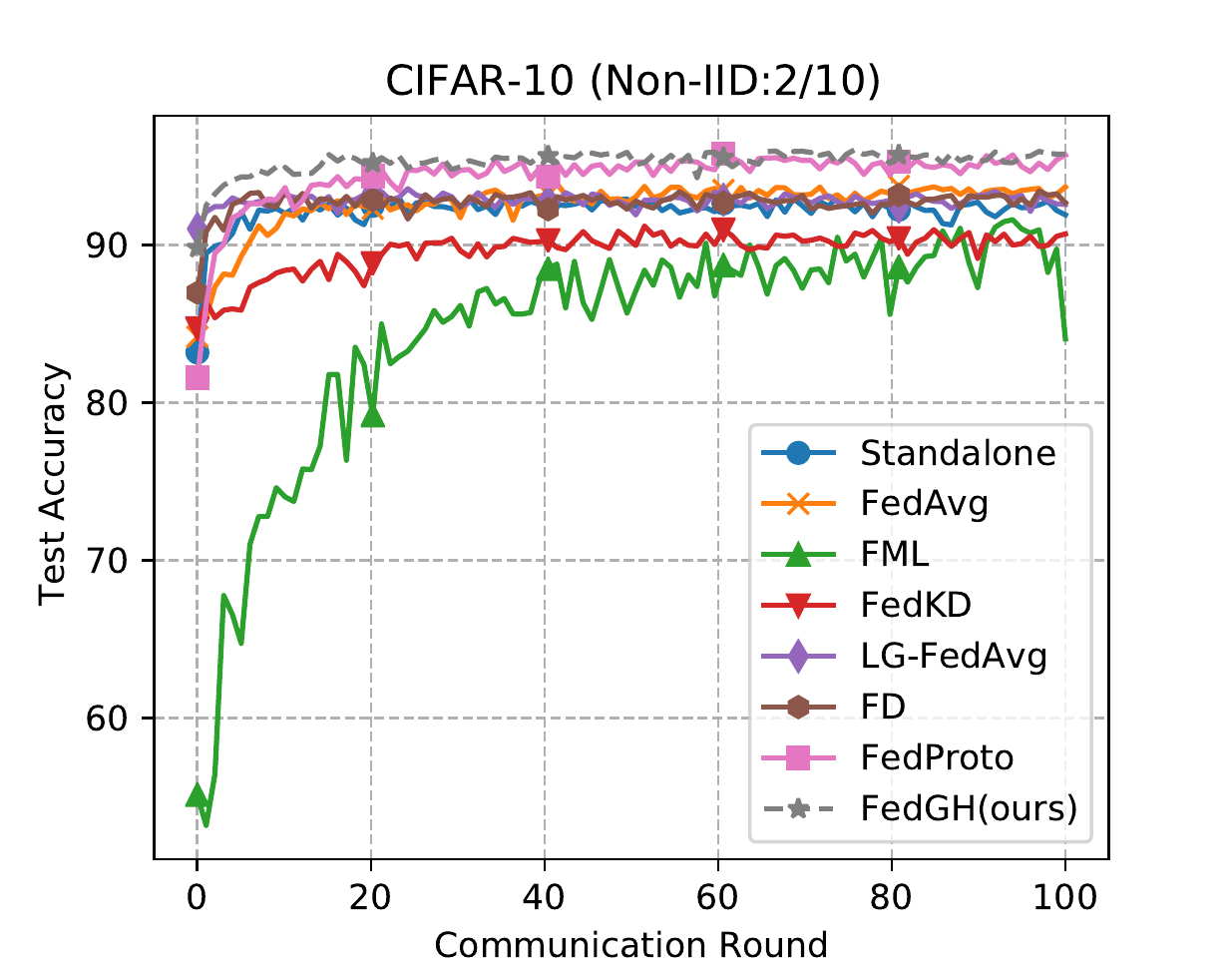}
\end{minipage}%
\begin{minipage}[t]{0.5\linewidth}
\centering
\includegraphics[width=1.8in]{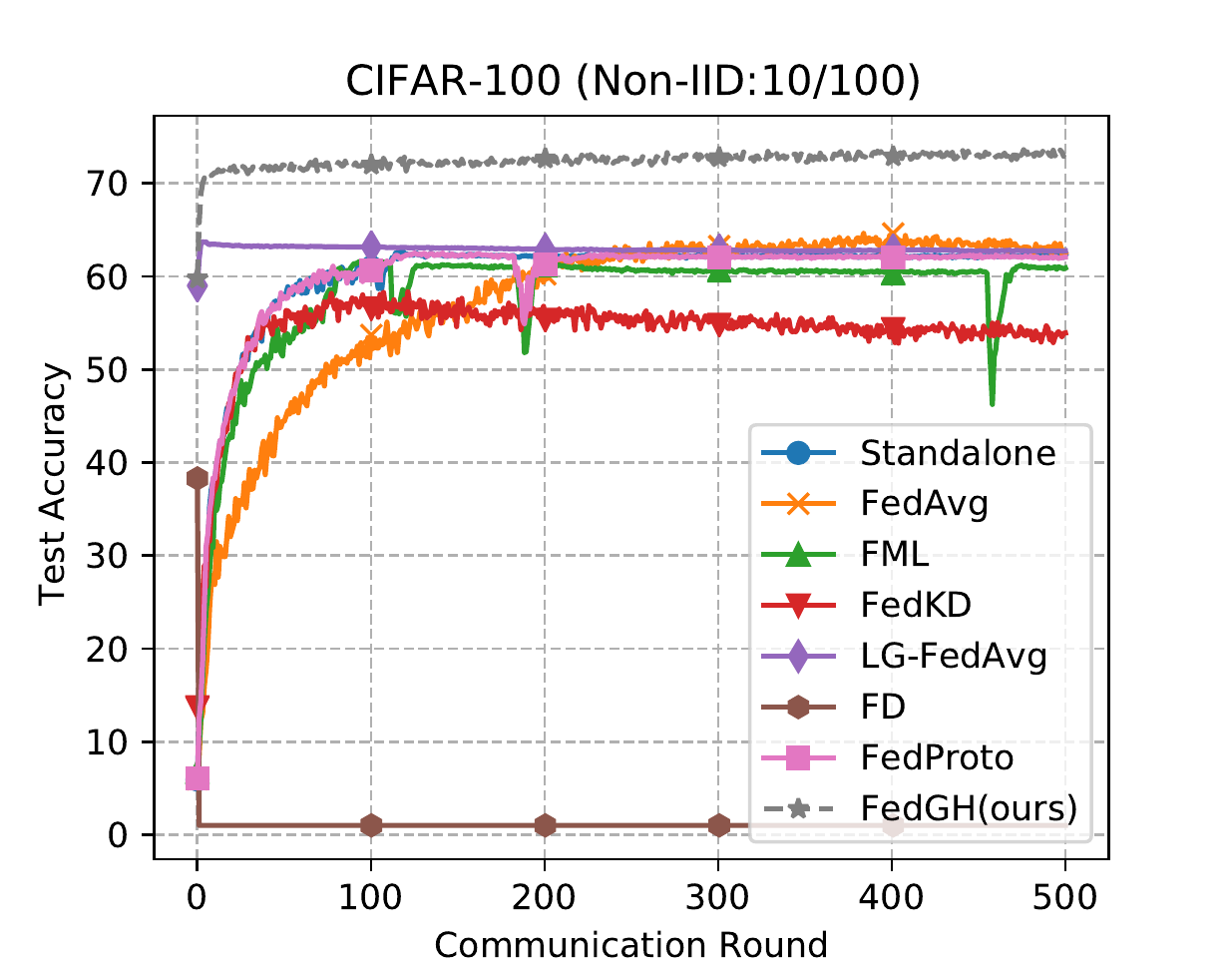}
\end{minipage}%
\vspace{-0.5em}
\caption{Test accuracy varies with the communication rounds when $N=10, C=100\%$.}
\label{fig:compare-N10}
\vspace{-1em}
\end{figure}

\subsubsection{Model-Homogeneity FL Setting}
To compare \methodname{} with baselines with different total numbers of clients $N$ and client participating rates $C$, we design three settings: $\{(N=10, C=100\%), (N=50, C=20\%), (N=100, C=10\%)\}$. For a fair comparison, we ensure that the number of clients participating in each round is the same (i.e., $K=N \times C = 10$). The results are illustrated in Tab.~\ref{tab:model-homo}.
It can be observed that \methodname{} consistently achieves the highest model accuracy across experimental conditions. On average, it outperforms the best baseline {\tt{FedProto}} by $0.54\%$ and $8.87\%$ under CIFAR-10 and CIFAR-100, respectively. Since most algorithms achieve high accuracy when the batch size is set to $512$ on CIFAR-10, the accuracy improvement of \methodname{} is still significant. In addition, the obvious accuracy improvement of \methodname{} on CIFAR-100 further demonstrates its effectiveness in tackling statistical heterogeneity (non-IID issue). Fig.~\ref{fig:compare-N10} shows that \methodname{} converges to the highest accuracy at the fastest rate, demonstrating its high efficiency.

\subsubsection{Model-Heterogeneity FL Setting}
In this setting, we vary the number of filters in the convolutional layers and the dimension of fully-connected layers in CNN model to obtain $5$ heterogeneous models: CNN-$\{1, 2, ..., 5\}$, 
the detailed model structures and sizes are reported in Tab.~\ref{tab:model-structures}. We distribute them evenly among the clients (it is still possible for different clients to have models with the same structure).
In {\tt{FML}} and {\tt{FedKD}}, we let CNN-$\{1, 2, ..., 5\}$ be clients' heterogeneous large models, and CNN-$5$ with the smallest model size be clients' homogeneous small models for aggregation at server. 

The results are shown in Tab.~\ref{tab:model-hetero}. It can be observed that \methodname{} consistently achieves the highest model accuracy. It outperforms the best baseline {\tt{FedProto}} by $1.17\%$ and $1.83\%$ under CIFAR-10 and CIFAR-100, respectively. 
Meanwhile, \methodname{} requires the fewest communication rounds to reach the target accuracy, thereby achieving convergence the fastest. It achieves moderate CO under CIFAR-10. However, under the more challenging CIFAR-100 dataset, it incurs the lowest CO, reducing it by 85.53\% compared to the best-performing baseline {\tt{FedProto}}.

{\ReviewThree{
Tab.~\ref{tab:model-hetero} also shows that {\tt{FML}} fails to converge and {\tt{FedKD}} converges with obviously lower accuracy. The reason for the results may be that training the heterogeneous large model and the homogeneous small model locally only requires the hard loss and distillation loss of the output logits of the two models in {\tt{FML}}, which incurs less information interaction between the two models. And in the initial training rounds, the immature shared homogeneous small model may hinder the convergence of the local heterogeneous large model. {\tt{FedKD}} designs an adaptive hidden loss of the two models’ hidden states and an adaptive mutual distillation loss based on {\tt{FML}}, the increase of interacted knowledge between the two models benefits their convergence.

}}

\begin{table}[t]
\centering
\caption{Structures of five heterogeneous CNN models. In the convolutional layers, the kernel size is $5 \times 5$, the number of filters is $16$ or $32$, and the dimensions of fc$3$ are consistent with the classes in CIFAR-10 or CIFAR-100 datasets.}
\resizebox{1\linewidth}{!}{%
\begin{tabular}{|l|c|c|c|c|c|}
\hline
layer name         & CNN-1    & CNN-2   & CNN-3   & CNN-4   & CNN-5   \\ \hline
conv1              & 5$\times$5, 16   & 5$\times$5, 16  & 5$\times$5, 16  & 5$\times$5, 16  & 5$\times$5, 16  \\
conv2              & 5$\times$5, 32   & 5$\times$5, 16  & 5$\times$5, 32  & 5$\times$5, 32  & 5$\times$5, 32  \\
fc1                & 2000     & 2000    & 1000    & 800     & 500     \\
fc2                & 500      & 500     & 500     & 500     & 500     \\
fc3                & 10/100   & 10/100  & 10/100  & 10/100  & 10/100  \\ \hline
model size & 10.00 MB & 6.92 MB & 5.04 MB & 3.81 MB & 2.55 MB \\ \hline
\end{tabular}%
}
\label{tab:model-structures}
\vspace{-1em}
\end{table}

\begin{table}[t]
\centering
\caption{Comparison of average test accuracy and communication overhead (CO) under the model-heterogeneous FL setting. CO/c/r denotes the CO per client per round. Rounds (X) denotes the number of training rounds required to reach target accuracy X, and CO is the total communication traffic consumed for the target accuracy. $N=10$ and $C=100\%$. ``-'' indicates that the algorithm fails to converge or reach the target accuracy.}
\resizebox{\linewidth}{!}{%
\begin{tabular}{|l|cccc|cccc|}
\hline
           & \multicolumn{4}{c|}{CIFAR-10 (non-IID: 2/10)}                                                                                                                                                                                                                                            & \multicolumn{4}{c|}{CIFAR-100 (non-IID: 10/100)}                                                                                                                                                                                                                                         \\ \cline{2-9} 
Method     & \multicolumn{1}{c|}{\begin{tabular}[c]{@{}c@{}}Acc\\ (\%)\end{tabular}} & \multicolumn{1}{c|}{\begin{tabular}[c]{@{}c@{}}CO/c/r\\ (KB)\end{tabular}} & \multicolumn{1}{c|}{\begin{tabular}[c]{@{}c@{}}Rounds\\  (90\%)\end{tabular}} & \begin{tabular}[c]{@{}c@{}}CO\\ (KB)\end{tabular} & \multicolumn{1}{c|}{\begin{tabular}[c]{@{}c@{}}Acc\\ (\%)\end{tabular}} & \multicolumn{1}{c|}{\begin{tabular}[c]{@{}c@{}}CO/c/r\\ (KB)\end{tabular}} & \multicolumn{1}{c|}{\begin{tabular}[c]{@{}c@{}}Rounds\\  (70\%)\end{tabular}} & \begin{tabular}[c]{@{}c@{}}CO\\ (MB)\end{tabular} \\ \hline
{\tt{Standalone}} & \multicolumn{1}{c|}{96.62}                                              & \multicolumn{1}{c|}{0}                                                     & \multicolumn{1}{c|}{0}                                                        & 0                                                 & \multicolumn{1}{c|}{72.34}                                              & \multicolumn{1}{c|}{0}                                                     & \multicolumn{1}{c|}{0}                                                        & 0                                                 \\
{\tt{FML}}        & \multicolumn{1}{c|}{-}                                                  & \multicolumn{1}{c|}{-}                                                     & \multicolumn{1}{c|}{-}                                                        & -                                                 & \multicolumn{1}{c|}{-}                                                  & \multicolumn{1}{c|}{-}                                                     & \multicolumn{1}{c|}{-}                                                        & -                                                 \\
{\tt{FedKD}}      & \multicolumn{1}{c|}{80.16}                                              & \multicolumn{1}{c|}{-}                                                     & \multicolumn{1}{c|}{-}                                                        & -                                                 & \multicolumn{1}{c|}{52.70}                                              & \multicolumn{1}{c|}{-}                                                     & \multicolumn{1}{c|}{-}                                                        & -                                                 \\
{\tt{LG-FedAvg}}  & \multicolumn{1}{c|}{96.37}                                              & \multicolumn{1}{c|}{39.14}                                                 & \multicolumn{1}{c|}{11}                                                       & 4305.47                                           & \multicolumn{1}{c|}{72.33}                                              & \multicolumn{1}{c|}{391.41}                                                & \multicolumn{1}{c|}{39}                                                       & 149.07                                            \\
{\tt{FD}}         & \multicolumn{1}{c|}{96.13}                                              & \multicolumn{1}{c|}{\textbf{0.16}}                                         & \multicolumn{1}{c|}{4}                                                        & \textbf{6.25}                                     & \multicolumn{1}{c|}{-}                                                  & \multicolumn{1}{c|}{-}                                                     & \multicolumn{1}{c|}{-}                                                        & -                                                 \\
{\tt{FedProto}}   & \multicolumn{1}{c|}{96.47}                                              & \multicolumn{1}{c|}{7.81}                                                  & \multicolumn{1}{c|}{4}                                                        & 312.50                                            & \multicolumn{1}{c|}{72.80}                                              & \multicolumn{1}{c|}{\textbf{39.06}}                                        & \multicolumn{1}{c|}{266}                                                      & 101.47                                            \\ \hline
\methodname{}      & \multicolumn{1}{c|}{\textbf{97.60}}                                     & \multicolumn{1}{c|}{23.45}                                                 & \multicolumn{1}{c|}{\textbf{2}}                                               & 468.91                                            & \multicolumn{1}{c|}{\textbf{74.13}}                                     & \multicolumn{1}{c|}{214.88}                                                & \multicolumn{1}{c|}{\textbf{7}}                                               & \textbf{14.69}                                    \\ \hline
\end{tabular}%
}
\label{tab:model-hetero}
\end{table}

\subsection{Case Studies}
In this section, we evaluate the robustness of the approaches to Non-IIDness and client participation rates, and we also test whether {\methodname{}} is sensitive to the only hyperparameter $\eta_\theta$ (the learning rate of the global prediction header).

\subsubsection{Robustness to Non-IIDness}
We test \methodname{} and state-of-the-art model-heterogeneous baselines: {\tt{LG-FedAvg}} and {\tt{FedProto}} on CIFAR-10 and CIFAR-100 with different Non-IID degrees. Specifically, we set $N=10$ and $C=100\%$. Then, we distribute $\{2, 4, 6, 8, 10\}$ classes of samples into each client under CIFAR-10, and we allocate $\{10, 30, 50, 70, 90, 100\}$ classes of samples into each client under CIFAR-100. The more classes of samples a client has, the lower the Non-IID degree.

\begin{figure}[t]
\centering
\begin{minipage}[t]{0.5\linewidth}
\centering
\includegraphics[width=1.8in]{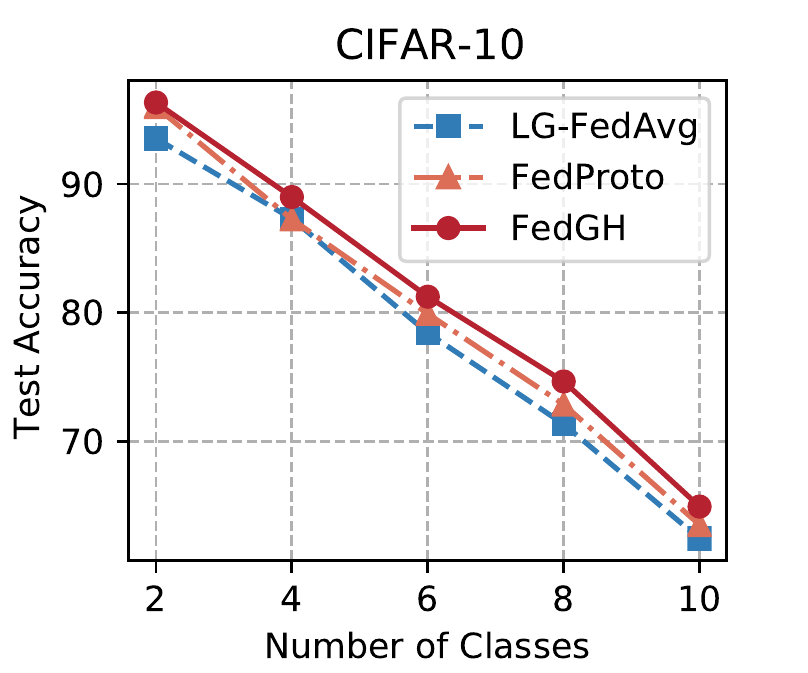}
\end{minipage}%
\begin{minipage}[t]{0.5\linewidth}
\centering
\includegraphics[width=1.8in]{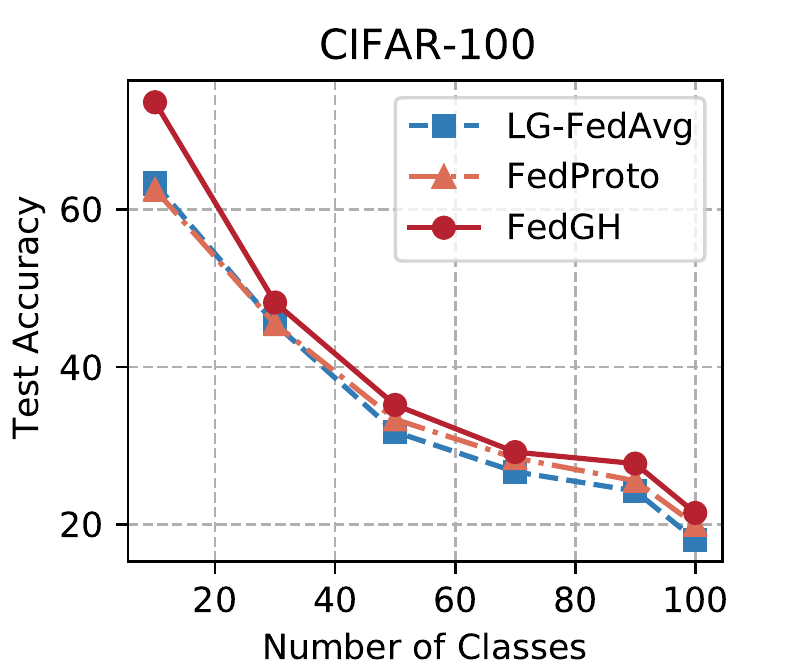}
\end{minipage}%
\vspace{-0.5em}
\caption{Robustness to Non-IIDness.}
\label{fig:case-noniid}
\vspace{-1em}
\end{figure}

Figure~\ref{fig:case-noniid} shows that \methodname{} consistently achieves the highest model accuracy across different Non-IID degrees on both CIFAR-10 and CIFAR-100, which demonstrates its robustness to Non-IIDness. In addition, it can also be observed that the model accuracy degrades as the number of classes increases (i.e., more IID) as personalization of local models is less advantageous as data heterogeneity decreases (which corroborates findings in \cite{FML}).

\subsubsection{Robustness to Partial Participation}
We test \methodname{} and state-of-the-art model-heterogeneous baselines: {\tt{LG-FedAvg}} and {\tt{FedProto}} on CIFAR-10 and CIFAR-100 with different client participation rates. Specifically, we set $N=100$ and vary $C \in \{0.1, 0.3, 0.5, 0.7, 0.9, 1\}$ under CIFAR-10 (Non-IID:2/10) and CIFAR-100 (Non-IID:10/100). 

\begin{figure}[t]
\centering
\begin{minipage}[t]{0.5\linewidth}
\centering
\includegraphics[width=1.8in]{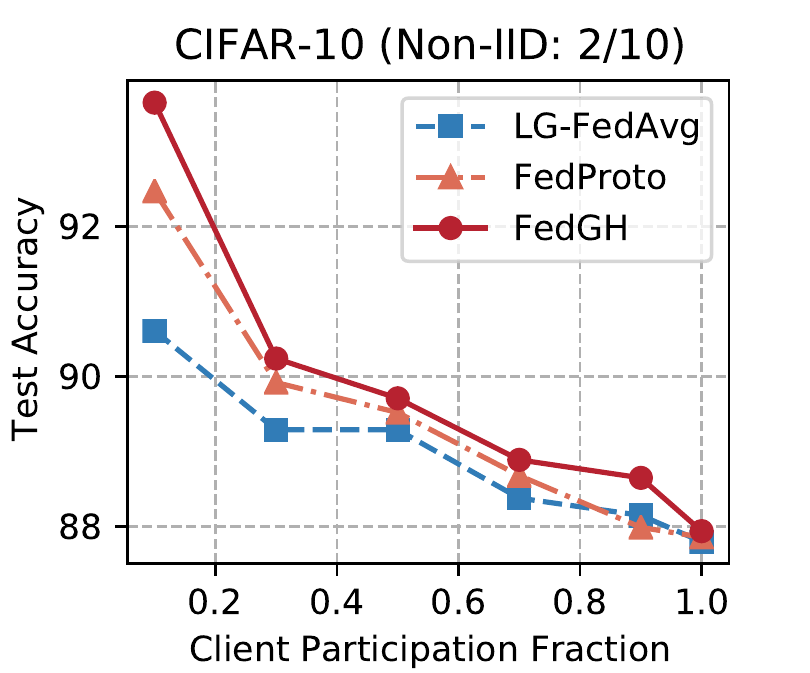}
\end{minipage}%
\begin{minipage}[t]{0.5\linewidth}
\centering
\includegraphics[width=1.8in]{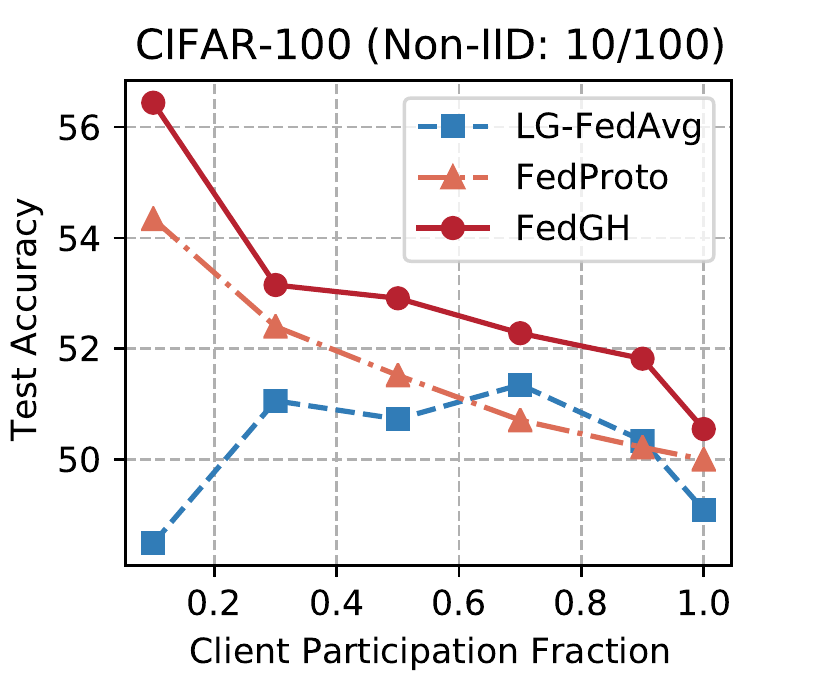}
\end{minipage}%
\vspace{-0.5em}
\caption{Robustness to client participation rate.}
\label{fig:case-frac}
\vspace{-1em}
\end{figure}

Figure~\ref{fig:case-frac} shows that \methodname{} consistently achieves the highest model accuracy under different client participation rates on both CIFAR-10 and CIFAR-100. This demonstrates its robustness to client participation rate. It can also be observed that the model accuracy decreases as the client participation rate increases. As more clients participate in one round of FL model training, generalization is enhanced but personalization becomes more challenging.

\subsubsection{Sensitivity to Hyperparameter $\eta_\theta$}
We test the sensitivity of {\methodname{}} to its only hyperparameter $\eta_\theta$ (the learning rate of the global prediction header on the server) on CIFAR-10 (Non-IID:2/10) and CIFAR-100 (Non-IID:10/100) datasets with the following settings: $N=10$, $C=100\%$, SGD optimizer with the global header's learning rate $\eta_\theta=\{0.001, 0.003, 0.01, 0.03, 0.1, 0.3, 1\}$ and the local model's learning rate $\eta_\omega=0.1$.

\begin{figure}[t]
\centering
\begin{minipage}[t]{0.5\linewidth}
\centering
\includegraphics[width=1.8in]{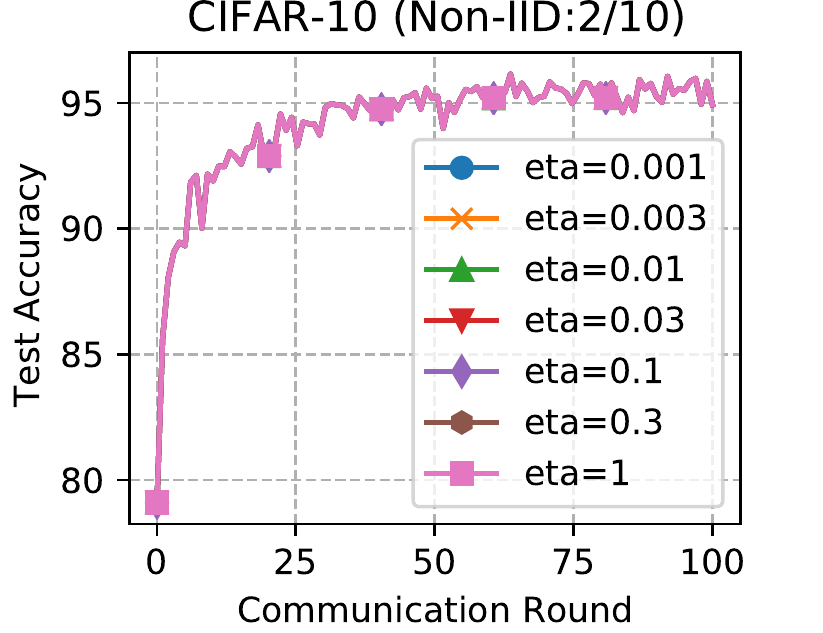}
\end{minipage}%
\begin{minipage}[t]{0.5\linewidth}
\centering
\includegraphics[width=1.8in]{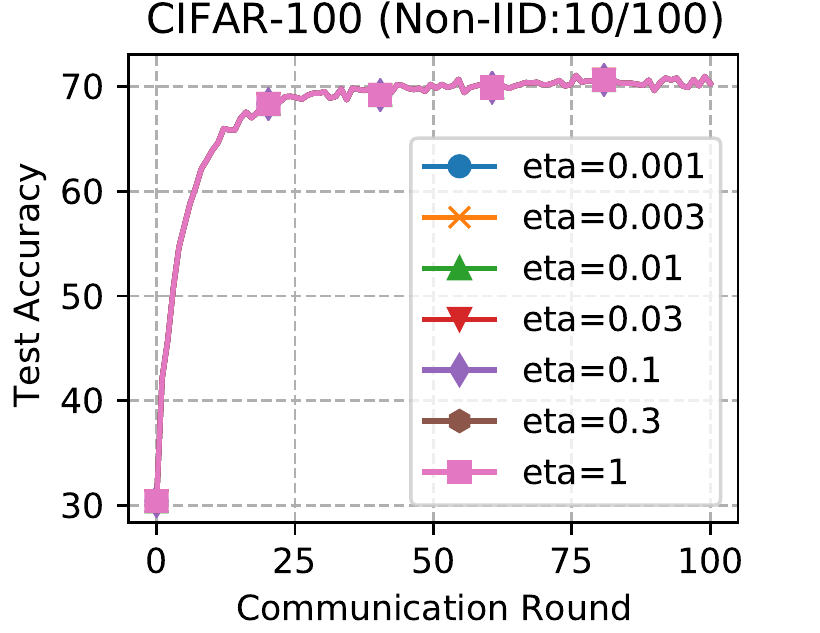}
\end{minipage}%
\vspace{-0.5em}
\caption{Sensitivity to the global header learning rate $\eta_\theta$.}
\label{fig:case-lr}
\vspace{-1em}
\end{figure}

Fig.~\ref{fig:case-lr} shows that the learning rate of the global prediction header has no influence on the performance of \methodname{}, indicating that \methodname{} is not sensitive to this hyperparameter. The reason is that there are few local average representations (LARs) from all client classes used for training the global prediction header. This training process is relatively easier than training local large complete models. Thus, the learning rate has no influence on it.

{\ReviewOne{\section{Conclusions and Future Work}}}
In this paper, we proposed a model-heterogeneous FL framework - \methodname{}. It utilizes the same-dimension representations extracted by clients' local heterogeneous feature extractors to train a homogeneous global prediction header shared by all clients, which can transfer all-class knowledge to clients by replacing clients' local headers. Theoretical derivations prove the non-convex convergence rate of \methodname{}. Extensive experiments demonstrate its superiority in terms of model performance and communication efficiency in both model-homogeneous and model-heterogeneous FL settings.

{\ReviewOne{
There are two promising directions in future work: a) since computing the local averaged representation (LAR) of each class for each client may incur information distortion, especially when one class has a large number of data samples, exploring an integrated representation containing as much local data information as possible benefits boosting the performance of the global classification header. b) Considering the fusion of the generalized global header and the personalized local header may improve the generalization and personalization of each client’s final classification header. \\
}}

\begin{acks}
This research is supported in part by the National Science Foundation of China under Grant 62272253, 62272252 and 62141412; the Fundamental Research Funds for the Central Universities; the National Research Foundation Singapore and DSO National Laboratories under the AI Singapore Programme (AISG Award No: AISG2-RP-2020-019); the RIE 2020 Advanced Manufacturing and Engineering (AME) Programmatic Fund (No. A20G8b0102), Singapore; the Joint NTU-WeBank Research Centre on Fintech (NWJ-2020-008); and the Nanyang Assistant Professorship (NAP). 
\end{acks}


\bibliographystyle{ACM-Reference-Format}
\balance
\bibliography{references}

\appendix
\onecolumn

\section{Proof for Lemma~\ref{lemma:AfterAggregation}}\label{sec:proof-Lemma2}
\begin{proof}
\begin{equation}\label{eq:15}
\begin{aligned}
 \mathcal{L}_{(t+1) E+0} & =\mathcal{L}_{(t+1) E}+\mathcal{L}_{(t+1) E+0}-\mathcal{L}_{(t+1) E}  \\
& \stackrel{(a)}{=} \mathcal{L}_{(t+1) E}+\mathcal{L}\left(\left(\varphi_k^{t+1}, \theta^{t+1}\right) ; \boldsymbol{x}, y\right)-\mathcal{L}\left(\left(\varphi_k^{t+1}, \theta_k^{t+1}\right) ; \boldsymbol{x}, y\right) \\
& \stackrel{(b)}{\leqslant} \mathcal{L}_{(t+1) E}+\left\langle\nabla \mathcal{L}\left(\left(\varphi_k^{t+1}, \theta_k^{t+1}\right)\right),\left(\left(\varphi_k^{t+1}, \theta^{t+1}\right)-\left(\varphi_k^{t+1}, \theta_k^{t+1}\right) \right) \rangle \right. 
 +\frac{L_1}{2}\left\|\left(\varphi_k^{t+1}, \theta^{t+1}\right)-\left(\varphi_k^{t+1}, \theta_k^{t+1}\right)\right\|_2^2  \\
& \stackrel{(c)}{\leqslant} \mathcal{L}_{(t+1) E}+\frac{L_1}{2}\left\|\left(\varphi_k^{t+1}, \theta^{t+1}\right)-\left(\varphi_k^{t+1}, \theta_k^{t+1}\right)\right\|_2^2  \\
& \stackrel{(d)}{\leqslant} \mathcal{L}_{(t+1) E}+\frac{L_1}{2}\left\|\theta^{t+1}-\theta_k^{t+1}\right\|_2^2  \\
& \stackrel{(e)}{=} \mathcal{L}_{(t+1) E}+\frac{L_1}{2}\left\|\theta^t-\eta \nabla \mathcal{L}\left(\theta^t\right)-\theta_k^t+\eta \nabla \mathcal{L}\left(\theta_k^t\right)\right\|_2^2  \\
& =\mathcal{L}_{(t+1) E}+\frac{L_1}{2}\left\|\theta^t-\theta_k^t+\eta\left(\nabla \mathcal{L}\left(\theta_k^t\right)-\nabla \mathcal{L}\left(\theta^t\right)\right)\right\|_2^2  \\
& \stackrel{(f)}{\leqslant} \mathcal{L}_{(t+1) E}+\frac{L_1}{2}\left\|\eta\left(\nabla \mathcal{L}\left(\theta_k^t\right)-\nabla \mathcal{L}\left(\theta^t\right)\right)\right\|_2^2  \\
& =\mathcal{L}_{(t+1) E}+\frac{\eta L_1}{2}\left\|\left(\nabla \mathcal{L}\left(\theta_k^t\right)-\nabla \mathcal{L}\left(\theta^t\right)\right)\right\|_2^2. 
\end{aligned}
\end{equation}

Take the expectation of $\mathcal{B}$ on both sides of Eq. \eqref{eq:15}, we have:
\begin{equation}\label{eq:16}
\begin{aligned}
 \mathbb{E}\left[\mathcal{L}_{(t+1) E+0}\right] & \leqslant \mathbb{E}\left[\mathcal{L}_{(t+1) E}\right]+\frac{\eta L_1}{2} \mathbb{E}\left[\left\|\left(\nabla \mathcal{L}\left(\theta_k^t\right)-\nabla \mathcal{L}\left(\theta^t\right)\right)\right\|_2^2\right] \\
& \stackrel{(g)}{\leqslant} \mathbb{E}\left[\mathcal{L}_{(t+1) E}\right]+\frac{\eta L_1 \delta^2}{2}.
\end{aligned}
\end{equation}

In Eq. \eqref{eq:15}, $(a)$: $\mathcal{L}_{(t+1) E+0}=\mathcal{L}\left(\left(\varphi_k^{t+1}, \theta^{t+1}\right); \boldsymbol{x}, y\right)$, i.e., at the start of the $(t+2)$-th round, the $k$-th client's local model is the combination of the local feature extractor $\varphi_k^{t+1}$ after local training in the $(t+1)$-th round, and the \textit{global} header $\theta^{t+1}$ after training in the $(t+1)$-th round. $\mathcal{L}_{(t+1) E}=\mathcal{L}\left(\left(\varphi_k^{t+1}, \theta_k^{t+1}\right) ; \boldsymbol{x}, y\right)$, i.e., in the $E$-th (last) local iteration of the $(t+1)$-th round, the $k$-th client's local model consists of the feature extractor $\varphi_k^{t+1}$ and the \textit{local} prediction header $\theta_k^{t+1}$. 
$(b)$ follows Assumption~\ref{assump:Lipschitz}. 
$(c)$: the inequality still holds when the second term is removed from the right-hand side. 
$(d)$: both $\left(\varphi_k^{t+1}, \theta^{t+1}\right)$ and $\left(\varphi_k^{t+1}, \theta_k^{t+1}\right)$ have the same $\varphi_k^{t+1}$, the inequality still holds after it is removed.
$(e)$: model training through gradient descent, i.e., $\theta^{t+1}=\theta^t-\eta \nabla \mathcal{L}\left(\theta^t\right), \theta_k^{t+1}=\theta_k^t-\eta \nabla \mathcal{L}\left(\theta_k^t\right)$. Here, we assume that both the learning rate for training local models and the learning rate for training the global prediction header are $\eta$. 
$(f)$: the inequality still holds after removing $\left\|\theta^t-\theta_k^t\right\|_2^2$ from the right hand side. 
$(g)$ follows Assumption~\ref{assump:header}.

\end{proof}

\section{Proof for Theorem~\ref{theorem:one-round}}\label{sec:proof-Theorem1}
\begin{proof} 
Substituting Lemma~\ref{lemma:LocalTraining} into the second term on the right hand side of Lemma~\ref{lemma:AfterAggregation}, can have:
\begin{equation}
\begin{aligned}
\mathbb{E}\left[\mathcal{L}_{(t+1) E+0}\right] & \leqslant \mathcal{L}_{t E+0}-\left(\eta-\frac{L_1 \eta^2}{2}\right) \sum_{e=0}^E\left\|\mathcal{L}_{t E+e}\right\|_2^2+\frac{L_1 E \eta^2}{2} \sigma^2+\frac{\eta L_1 \delta^2}{2} \\
& \leqslant \mathcal{L}_{t E+0}-\left(\eta-\frac{L_1 \eta^2}{2}\right) \sum_{e=0}^E\left\|\mathcal{L}_{t E+e}\right\|_2^2+\frac{\eta L_1\left(E \eta \sigma^2+\delta^2\right)}{2}
\end{aligned}
\end{equation}
\end{proof}

\section{Proof for Theorem~\ref{theorem:non-convex}}\label{sec:proof-Theorem2}
\begin{proof} 
Theorem~\ref{theorem:one-round} can be re-expressed as:
\begin{equation}\label{eq:18}
\sum_{e=0}^E\left\|\mathcal{L}_{t E+e}\right\|_2^2 \leqslant \frac{\mathcal{L}_{t E+0}-\mathbb{E}\left[\mathcal{L}_{(t+1) E+0}\right]+\frac{\eta L_1\left(E \eta \sigma^2+\delta^2\right)}{2}}{\eta-\frac{L_1 \eta^2}{2}}.
\end{equation}

Take expectations of model $\omega$ on both sides of Eq. \eqref{eq:18}, we have:
\begin{equation}\label{eq:19}
\sum_{e=0}^E \mathbb{E}\left[\left\|\mathcal{L}_{t E+e}\right\|_2^2\right] \leqslant \frac{\mathbb{E}\left[\mathcal{L}_{t E+0}\right]-\mathbb{E}\left[\mathcal{L}_{(t+1) E+0}\right]+\frac{\eta L_1\left(E \eta \sigma^2+\delta^2\right)}{2}}{\eta-\frac{L_1 \eta^2}{2}} .
\end{equation}

Summing both sides of Eq. \eqref{eq:19} over $T$ rounds (i.e., $t \in [0,T-1]$) yields:
\begin{equation}
\frac{1}{T} \sum_{t=0}^{T-1} \sum_{e=0}^E \mathbb{E}\left[\left\|\mathcal{L}_{t E+e}\right\|_2^2\right] \leqslant \frac{\frac{1}{T} \sum_{t=0}^{T-1}\left(\mathbb{E}\left[\mathcal{L}_{t E+0}\right]-\mathbb{E}\left[\mathcal{L}_{(t+1) E+0}\right]\right)+\frac{\eta L_1\left(E \eta \sigma^2+\delta^2\right)}{2}}{\eta-\frac{L_1 \eta^2}{2}}.
\end{equation}

Since $\sum_{t=0}^{T-1}\left(\mathbb{E}\left[\mathcal{L}_{t E+0}\right]-\mathbb{E}\left[\mathcal{L}_{(t+1) E+0}\right]\right) \leqslant \mathcal{L}_{t=0}-\mathcal{L}^*$, we have:
\begin{equation} \label{eq:T-rounds}
\begin{aligned}
 \frac{1}{T} \sum_{t=0}^{T-1} \sum_{e=0}^E \mathbb{E}\left[\left\|\mathcal{L}_{t E+e}\right\|_2^2\right] & \leqslant \frac{\frac{1}{T}\left(\mathcal{L}_{t=0}-\mathcal{L}^*\right)+\frac{\eta L_1\left(E \eta \sigma^2+\delta^2\right)}{2}}{\eta-\frac{L_1 \eta^2}{2}} \\
& =\frac{2\left(\mathcal{L}_{t=0}-\mathcal{L}^*\right)+\eta L_1 T\left(E \eta \sigma^2+\delta^2\right)}{T\left(2 \eta-L_1 \eta^2\right)}  \\
& =\frac{2\left(\mathcal{L}_{t=0}-\mathcal{L}^*\right)}{T \eta\left(2-L_1 \eta\right)}+\frac{L_1\left(E \eta \sigma^2+\delta^2\right)}{2-L_1 \eta}.
\end{aligned}
\end{equation}

If the local model can converge, the above equation satisfies
\begin{equation}
\frac{2\left(\mathcal{L}_{t=0}-\mathcal{L}^*\right)}{\operatorname{T\eta }\left(2-L_1 \eta\right)}+\frac{L_1\left(E \eta \sigma^2+\delta^2\right)}{2-L_1 \eta} \leqslant \epsilon.
\end{equation}
Then, we can obtain:
\begin{equation}
T \geqslant \frac{2\left(\mathcal{L}_{t=0}-\mathcal{L}^*\right)}{\eta \epsilon\left(2-L_1 \eta\right)-\eta L_1\left(E \eta \sigma^2+\delta^2\right)}.
\end{equation}

Since $T>0, \mathcal{L}_{t=0}-\mathcal{L}^*>0$, we can further derive:
\begin{equation}
\eta \epsilon\left(2-L_1 \eta\right)-\eta L_1\left(E \eta \sigma^2+\delta^2\right) > 0,    
\end{equation}
i.e.,
\begin{equation}\label{eq:24}
\eta < \frac{2 \epsilon-L_1 \delta^2}{L_1\left(\epsilon+E \sigma^2\right)}.
\end{equation}

The right-hand side of Eq. \eqref{eq:24} are all constants. Thus, the learning rate $\eta$ is upper bounded. When $\eta$ satisfies the above condition, the second term of the right-hand side of Eq.~\eqref{eq:T-rounds} is a constant. It can be observed from the first term of Eq.~\eqref{eq:T-rounds} the non-convex convergence rate satisfies $\epsilon \sim \mathcal{O}\left(\frac{1}{T}\right)$.
\end{proof}

\end{document}